\documentclass{article}
\usepackage[hyphens,spaces,obeyspaces]{url}
\usepackage{spconf,amsmath,graphicx}
\usepackage{color}

\usepackage{siunitx}

\usepackage{lscape}
\usepackage{rotating}
\usepackage{afterpage}
\usepackage{caption}
\usepackage{makecell}
\usepackage{csquotes}
\usepackage{units}
\usepackage{xspace}
\usepackage{hyperref}

\usepackage{booktabs}
\usepackage{array}
\usepackage{bm}
\usepackage{tipa}
\usepackage{tikzit}
\usetikzlibrary{arrows.meta}

\tikzstyle{medium box}=[fill=white, draw=black, shape=rectangle, minimum width=2cm, minimum height=.5]
\tikzstyle{GRASS}=[fill=black, draw=black, shape=rectangle]
\tikzstyle{GRASSdots}=[fill=none, draw=none, shape=circle]
\tikzstyle{Module}=[fill=white, draw=black, shape=circle, minimum width=2cm, minimum height=1cm]
\tikzstyle{Module2}=[fill=white, draw=black, shape=rectangle, minimum width=2cm, minimum height=1cm]

\tikzstyle{dashed arrow right}=[->, dashed]
\tikzstyle{arrow right}=[->]
\tikzstyle{big arrow right}=[-{Triangle[width=10pt,length=8pt]}, line width=6pt]
\tikzstyle{super big arrow right}=[-{Triangle[width=23pt,length=13pt]}, line width=15pt]
\usepackage{forest}
\forestset{qtree/.style={for tree={parent anchor=south, 
           child anchor=north,align=center,inner sep=0pt}}}
\usepackage{flexisym}
\usetikzlibrary{arrows.meta,bending,positioning}
\newcommand\ppbb{path picture bounding box}
  
\tikzset{switch/.style = {minimum size=1.7em,
                 path picture={  \draw
                 ([xshift=-2mm]  \ppbb.center)  -- ++ (30:30mm);
                                 \draw[shorten >=5mm,-]
                 (\ppbb.west) edge (\ppbb.center)
                 (\ppbb.east)  --  (\ppbb.center);
                                 \draw[<->]
                 ([yshift=-2mm] \ppbb.center) arc[start angle=-15, end angle=75, radius=5.5mm];   
                            },
                 label={[yshift=1mm] above:#1},
                 node contents={}},}

\newcommand{\GRCS}{GRASS\_CS\xspace}
\newcommand{\KICS}{KIEL\_CS\xspace}
\newcommand{\GRRS}{GRASS\_RS\xspace}

\urldef{\wikiurl}\url{https://dumps.wikimedia.org/dewiki/20220701}

\newcommand{\grass}{\textit{GRASS} corpus\xspace}
\newcommand*\rot{\rotatebox{90}}
\title{Using Kaldi for automatic speech recognition of conversational Austrian German}
\name{{Julian Linke, Saskia Wepner, Gernot Kubin, Barbara Schuppler}}
\address{
	Signal Processing and Speech Communication Laboratory\\Graz University of Technology, Inffeldgasse 16c, 8010 Graz, Austria\\
}

\begin{document}
%
\maketitle
\begin{abstract} 
As dialogue systems are becoming more and more interactional and social, also the accurate automatic speech recognition (ASR) of conversational speech is of increasing importance. This shifts the focus from short, spontaneous, task-oriented dialogues to the much higher complexity of casual face-to-face conversations. However, the collection and annotation of such conversations is a time-consuming process and data is sparse for this specific speaking style. This paper presents ASR experiments with read and conversational Austrian German as target.  In order to deal with having only limited resources available for conversational  German and, at the same time, with a large variation among speakers with respect to pronunciation characteristics, we improve a Kaldi-based ASR system by incorporating a (large) knowledge-based pronunciation lexicon, while exploring different data-based methods to restrict the number of pronunciation variants for each lexical entry. We achieve best WER of $\unit[0.4]{\%}$ on Austrian German read speech and best average WER of $\unit[48.5]{\%}$ on conversational speech. We find that by using our best pronunciation lexicon a similarly high performance can be achieved than by increasing the size of the data used for the language model by approx. 360\% to 760\%. Our findings indicate that for low-resource scenarios -- despite the general trend in speech technology towards using data-based methods only -- knowledge-based approaches are a successful, efficient method. 

\end{abstract}
\section{Introduction}
\label{sec:intro}

Automatic Speech Recognition (ASR) for conversational speech (CS) has become more relevant in recent years. On the one hand, from a speech technology point of view, there is a growing interest in social robots that perform human-machine interaction. Speech agents that are supposed to act as conversation partners are only useful if they are actually capable of having a fluent, natural conversation in terms of managing timing, speaker disfluencies or decent hypothesis re-ranking \cite{Baumann}. On the other hand, from a speech science point of view, also the phonetic and psycholinguistic analysis of ever-day speech processes has received more and more attention \cite{Tucker2016, Wagner2015}. Given the higher variation in spontaneous speech compared to the speech occurring in controlled production experiments, the analysis of this speaking style also requires larger amounts of speech data, transcribed ideally automatically by means of an ASR system (e.g., \cite{RR2017}). However, automatic transcriptions of conversational speech still differ widely from those by human annotators \cite{lopez2022evaluation}. Despite this interest in conversational speech from both speech science and technology, most state-of-the-art ASR systems still continue to be developed for speaking styles that are neither fully spontaneous nor interactive. One of the reasons for this phenomenon is the lack of large enough (transcribed or raw) speech resources for more spontaneous speaking styles. In this paper, we face the challenge of ASR for casual, conversational speech, for a low-resourced language variety (Austrian German). 


\subsection{Categorization of spontaneous speech corpora}

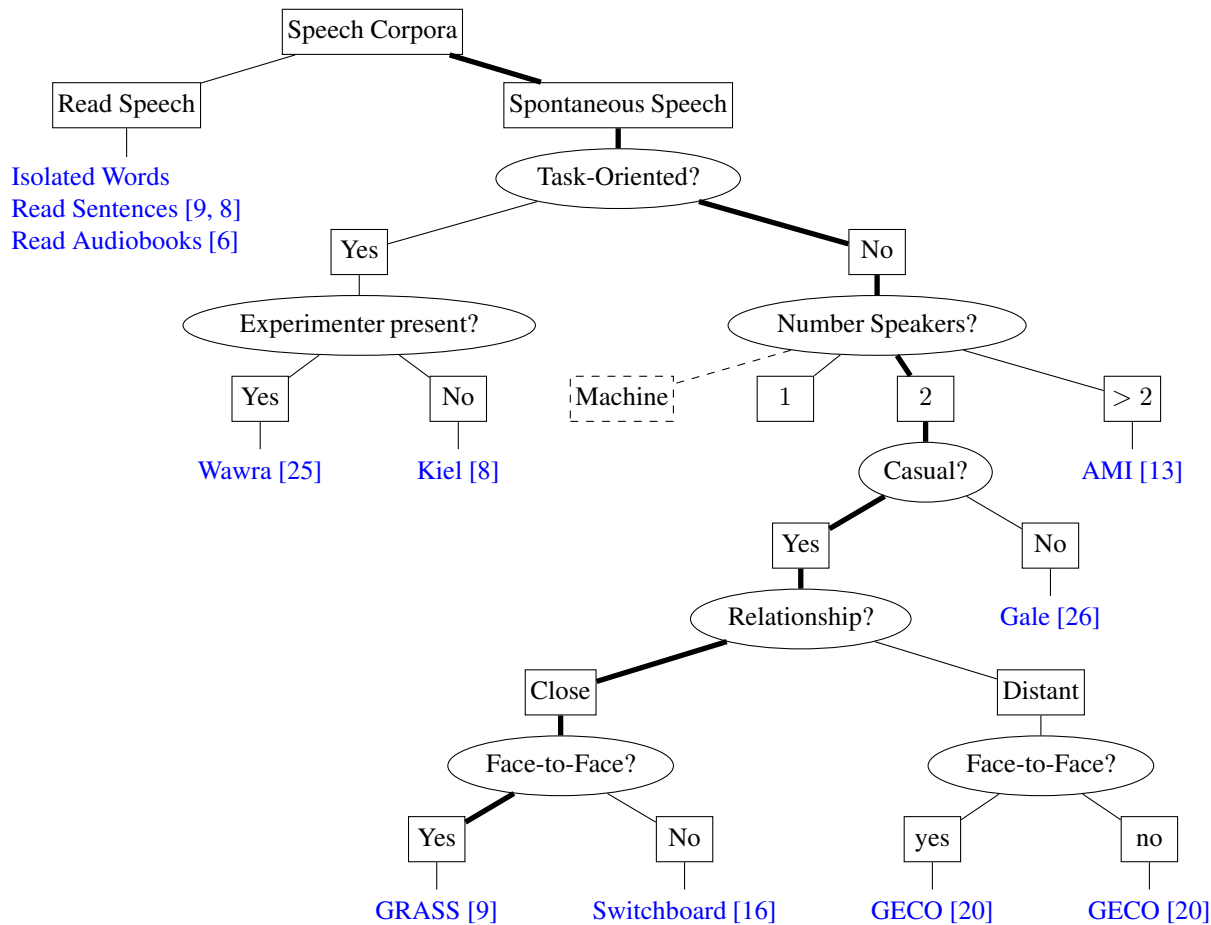
\begin{figure*}[ht!]
    \forestset{
            b/.style={draw=none,blue},
            r/.style={draw,ellipse},
            m/.style={dashed,edge=dashed},
            g/.style={edge={line width=2pt}}
    }
    \centering
    \begin{forest}for tree={
            align=left,
            draw, 
            inner sep=2pt,
            l sep=.25cm, 
            s sep=1.1cm, 
            minimum height=.6cm,
            minimum width=.75cm},
[Speech Corpora
    [Read Speech
        [Isolated Words\\Read Sentences \cite{schuppler2014grass,Kohler2017}\\Read Audiobooks \cite{Panayotov2015}, b]
    ]
    [Spontaneous Speech, g
        [Task-Oriented?,r,g
            [Yes
                [Experimenter present?,r
            [Yes
                [Wawra \cite{2015Wawra},b]
            ]
            [No
                [Kiel \cite{Kohler2017},b]
            ]
        ]
            ]
            [No, g
                [Number Speakers?,r,g
                    [Machine, m,
                    ]
                    [$1$
                    ]
                    [$2$, g
                        [Casual?, r,g
                            [Yes, g
                                [Relationship?,r,g
                                    [Close, g
                                        [Face-to-Face?,r,g
                                            [Yes, g
                                                [GRASS \cite{schuppler2014grass},b]
                                            ]
                                            [No
                                                [Switchboard \cite{godfrey1992switchboard},b]
                                            ]
                                        ]
                                    ]
                                    [Distant
                                         [Face-to-Face?,r
                                            [yes
                                                [GECO \cite{Schweitzer2015AttentionPE},b]
                                            ]
                                            [no
                                                [GECO \cite{Schweitzer2015AttentionPE},b]
                                            ]         
                                        ]
                                    ]
                                ]
                            ]
                            [No
                                [Gale \cite{2013GALE},b]
                            ]
                        ] 
                    ]
                    [$>2$
                        [AMI \cite{2006AMI},b]
                    ]
                ]
            ]
        ]
    ]
]
\end{forest}
    \caption{Categorization of speech corpora for  different speaking styles. The tree structure is defined by statements (black rectangles), questions (ellipses) and chosen examples (blue). The dashed rectangle indicates the possibility of machine-oriented interaction between speakers and machines (i.e., dialogue systems). The bold line indicates the path for \GRCS.}
    \label{fig:CSstyles}
\end{figure*}

In ASR literature, speaking styles are defined after different criteria, and terms for corpus categorization such as "read", "spontaneous" and "conversational" may actually point to corpora of very different characteristics. Figure \ref{fig:CSstyles} shows a categorization scheme that helps us describing the style of the corpora we use in this paper, and in general, helps us defining which of the corpora widely used in the ASR community are actually comparable to each other. Note that we do not present a full categorization of all possibilities in Figure \ref{fig:CSstyles}, but end the tree at those points, where the corpora used in this paper drop out. 
\par
In general, we can distinguish read from spontaneous speech, where spontaneous contrasts from read given that lexemes and their word order are planned spontaneously. Examples for read speech are LibriSpeech \cite{Panayotov2015}, for which state-of-the-art speech recognition systems reach a performance of $1.4\%$ Word Error Rate (WER) \cite{2020Zhang}. Also the read speech components of the Kiel \cite{Kohler2017} and GRASS corpus \cite{schuppler2014grass} used in this paper fall into this category (for more detail see Section \ref{sec:materials}).  
\par
Next, we distinguish spontaneous speech with respect to whether it is task-oriented or not. Speech from task-oriented dialogues are in general characterized by covering a specific domain, and that speakers chat less freely than when the topic of a conversation is open. Task-oriented dialogues may further be distinguished with respect to whether an experimenter is present or not. We define as experimenter a person present in the recordings that guides the conversation. That person could be a linguist or a  professionally trained broadcast interviewer. This type of speech is characterized by utterances of relatively complete syntactic structures which are pronounced carefully, given that trained speakers are involved. In contrast, in casual conversations, speakers strongly reduce their pronunciation and produce syntactically incomplete structures \cite{johnson2004massive}. An example for task-oriented dialogues without experimenter present are the dialogues in Verbmobil \cite{wahlster1993verbmobil} and in the Kiel Corpus \cite{Kohler2017}. These short dialogues last for approx. 2 - 20 minutes each, a time span that does not allow the speakers to \textit{forget about the recording situation}, which affects the naturalness of the resulting speaking style. 
\par
Another way how to categorize speech corpora is with respect to the (number of) interlocutor(s). We distinguish (spontaneous) monologues and machine-oriented dialogues that both do not show cross-talk, from conversations between two or more speakers (e.g., conversations between three speakers in a casual setting \cite{torreira:hal-00608402}; between even more speakers in a meeting setting \cite{2006AMI}). With increasing number of speakers, the challenge for ASR is rising, as one needs to deal with overlapping speech, which comes not only with acoustic difficulties, but also with structural speech phenomena such as co-completion, turn-competition and broken turns. For the AMI-Meeting corpus, WERs of approx. 21.2\% have been achieved \cite{kanda21_interspeech}.  
\par
When focusing on conversations between two speakers, we may further distinguish whether the topic of the conversation is casual or professional, as we assume that casual topics also lead to a speaking style that is characterized by more pronunciation variation and/or a stronger use of dialectal variants. Pronunciation variation may become even more salient when speakers have a close relationship to each other (and are maybe even from the same dialectal area). All of these effects on style may be continuous in some language areas (e.g., in Austrian German), or diglossic in others (e.g., in Swiss German) \cite{Stpkowska2012DiglossiaAC}. We are aware of the sociolinguistic fact that the effect of relationship on the speaking style is not comparable across languages. 
\par
Finally, we categorize conversational speech  with respect to whether they occurred face-to-face or not. In the widely used Switchboard corpus \cite{godfrey1992switchboard}, speakers who knew each other well were having telephone conversations, where the speakers were not able to benefit from visual cues and needed to deal with reduced sound quality, forcing them to pronounce more clearly and to avoid overlapping talk. ASR results for Switchboard are in the range of $\unit[4.3]{\%}$ to $\unit[5.1]{\%}$ WER \cite{tuske21_interspeech, xiong2018microsoft}.
\par
The IMS GECO database from Stuttgart contains conversations between two speakers of a distant relationship, in face-to-face (GECO\_Multi) and in a non-face-to-face setting where speakers were separated by a solid wall (GECO\_Mono) \cite{schweitzer13_interspeech,Schweitzer2015AttentionPE}. First word recognition results with GECO correctly identified only $\unit[25]{\%}$ of the words \cite{arnold2017words}.
\par 
The corpus in focus of this paper is the conversational component of the GRASS corpus \cite{schuppler2017corpus}, containing topic-open, casual, face-to-face conversations between two closely related persons that last for one full hour, with no experimenter present. So far, there is little  data available for this specific speaking style. The Japanese \textit{Corpus of Everyday Japanese Conversations} (CEJC) \cite{koiso2018construction} is currently in development and will i.a. include recordings from an individual-based recording method. The material will involve recordings from 40 informants who are balanced in terms of sex and age while each recording will originate from a portable recording device by collecting 15h of speech data over a period of approx. two to three months in various everyday situations. To the best of our knowledge, so far there have not yet been published any  ASR experiments with CEJC; a study on dialogue situation recognition using CEJC showed that the system did not reach the level of human evaluation results \cite{chiba21interspeech}.
In summary, CEJC and the data used in this study (\GRCS) contain a broad variety of challenges resulting from speaker interaction in conversational speech. 


\subsection{Aims of this paper}

The main aim of this paper is to present ASR experiments for GRASS\_CS, a corpus of conversational Austrian German of a total of 30 hours of speech. There exists no other German database of comparable style from the same language variety. For ASR, we use a DNN-HMM Kaldi model \cite{Povey2011Kaldi, 2014Povey} and test whether the use of other German corpora for acoustic- and language modeling yields higher performances. Given the large difference in pronunciation between German and Austrian German, we investigate to which degree the integration of linguistic knowledge into the pronunciation lexicon improves ASR performance. 
\par
When dealing with conversational Austrian German, one might ask whether the difficulty for ASR lies in the language variety (i.e., German vs. Austrian German) or in the speaking style. To tear these effects apart, we also report ASR results for read Austrian German, and show that with a rather simple Kaldi recipe already state-of-the art results can be achieved.

\section{Materials}
\label{sec:materials}

\subsection{GRASS corpus}
The Graz corpus of Read And Spontaneous Speech (\grass) \cite{schuppler2014grass, schuppler2017corpus} contains about 30h of Austrian German read (RS) and conversational speech (CS) from 38 Austrian speakers (19f/19m). As language usage in CS varies strongly with educational level, social background and dialect region, speakers were selected who were born in the same broad dialect region (Eastern Austria), had been living in an urban area for years and had a higher education degree. For the CS component, 19 pairs of speakers who have been knowing each other for several years were recorded for one hour each without interruption in order to encourage a fluent, spontaneous conversation. There was no experimenter present in the recording room and there was no restriction in terms of chosen topic or speaking behavior, leading to the use of natural, partly dialectal pronunciation with typical characteristics such as frequently occurring overlapping speech, laughter, and the use of swear words \cite{schuppler2017corpus}. Despite the speakers' awareness of being recorded, they appeared to completely forget about the studio recording situation after a period of five to ten minutes, entering a casual conversation. Only after the hour of CS, speakers read short stories as well as selected isolated sentences. Both, RS and CS component were produced by the same speakers.

\subsection{Kiel corpus}
The Kiel Corpus of Spoken German \cite{Kohler2017} contains a total of $5\si{\hour}$ of read and spontaneous speech produced by speakers mainly coming from Northern Germany. Two spontaneous components are available: (1) the \enquote{appointment-making-scenario} part, which contains approx.~$4\si{\hour}$ of dialogues from $43$ speakers (22f/21m) who were making appointments. In this scenario, speech was only recorded if participants were holding a button pressed which was at the same time blocking the interlocutor's channel. Thus, this scenario effectively avoids overlapping speech. (2) the \enquote{video-task-scenario} part contains approx. $1\si{\hour}$ of dyadic conversations. In this scenario, manipulated video materials from a television series were presented separately to two participants who had the task to find the differences in the video. We used the spontaneous speech component from the Kiel Corpus (KICS) for our experiments with Austrian CS.

\subsection{IMS GECO database}
The IMS GECO database (GECO) \cite{schweitzer13_interspeech,Schweitzer2015AttentionPE} contains $46$ spontaneous dialogues of approx.~$25$ minutes between unfamiliar female speakers from two settings: 1) a unimodal setting with $22$ dialogues (GECO-Mono), where participants could not see each other because they were separated by a solid wall and 2) a multimodal setting with $24$ dialogues (GECO-Multi), with face-to-face conversations comparable to \GRCS. The unimodal setting involves $12$ speakers, where $7$ returned for the multimodal setting meaning that some dialogue pairs are present in both GECO-Mono and GECO-Multi. In both settings, speakers were free to choose the topics they wanted to discuss.

\begin{table}[t!]
 \caption{Overview of used data sets: German and Austrian German corpora, containing read (RS) and conversational speech (CS).}
 \label{tab:data}
 \centering
 \begin{tabular}{ l c r c c }
  \toprule
   \multicolumn{1}{l}{\textbf{Abbr.}} &
  \multicolumn{1}{c}{\textbf{\makecell{Variety}}} &\multicolumn{1}{c}{\textbf{Hours}} \\
  \midrule
   \GRCS     & Austrian German  & $14.3$~~~   \\
   \GRRS      & Austrian German  & $4.6$~~~    \\
   GECO      & German  & $18.7$~~~   \\
  \KICS     & German  & $5$~~~   \\ 
  \bottomrule
 \end{tabular}
\end{table}

\section{ASR for Read Speech}
\label{sec:exp1}

\subsection{Methods}
\label{sec:methods_RS}
Acoustic Models (AM) and Language Models (LM) were trained on data from the RS corpus. The \GRRS data set comprises 6h of speech, where each speaker reads mostly the same, phonetically balanced sentences. The training set included 33 speakers (5.25h), the validation set 2 speakers (0.37h) and the test set also 2 speakers (0.34h). We excluded one speaker, because of an atypically disfluent reading style.
\par
We extracted 13-dimensional MFCCs and performed cepstral mean and variance normalization (CMVN) while comparing a combination of different frame lengths $\{20\si{\milli\second}$, $25\si{\milli\second}$, $30\si{\milli\second}\}$ and frame shifts $\{\unit[7.5]{ms}, \unit[10]{ms}, \unit[12.5]{ms}\}$. For the acoustic models (AM), the initial diagonal GMM-HMM models (short GMM) comprise basic monophone and triphone training with MFCCs+$\Delta$+$\Delta \Delta$ features. 
\par
The lexicon was built with a G2P online tool \cite{reichel2012perma} for standard German. As this resource is not available for the Austrian variety of German, we applied a set of rules on phone-level to adapt its output towards standard Austrian German pronunciation \cite{schuppler2014pronunciation}, and phonological reduction phenomena, such as schwa-deletion. We reduced the phone set yielded by G2P in order to improve recognition performance using three rules: (R1), a replacement rule to devoice all alveolar and postalveolar fricatives and affricates (a common phonological process in standard Austrian German); (R2), a rule to split all diphthongs into two separate phones; (R3), a rule to unite short and long vowels, based on phonetic studies on Austrian German \cite{Moosmueller2007}. In total, we reduced the phone set from initially 64 to 38 phones. 


We used the SRILM toolkit with a Witten-Bell discounting for an N-gram language model (LM) of different orders \cite{Stolcke02}. The LM was generated given the text of all utterances from the entire RS component (\emph{train}, \emph{valid} and \emph{test}) since in the RS component all speakers read the same text. 

\subsection{Results}
\label{sec:RS_results}
Table \ref{tab:resultsRS} shows a summary of the results for RS. First, we analyzed the influence of different frame shifts and frame lengths with a trigram LM. 
Our experiments showed that different frame lengths of $f_{\mathrm{len}}=\{20\si{\milli\second}, 25\si{\milli\second}, 30\si{\milli\second}\}$ have less impact on the WERs than frame shifts $f_{\mathrm{sh}}=\{\unit[7.5]{ms}, \unit[10]{ms}, \unit[12.5]{ms}\}$. The best triphone WER ($\unit[0.96]{\%}$) was achieved with $f_{\mathrm{sh}}=\unit[12.5]{ms}$ and $f_{\mathrm{len}}=\unit[20]{ms}$. Monophone and triphone models performed similarly most of the time. Yet, with a frame shift of $f_{\mathrm{sh}}=\unit[10]{ms}$ combined with frame lengths of $f_{\mathrm{len}}=\{\unit[25]{ms}, \unit[30]{ms}\}$, triphone models returned worse results. 
In order to further optimize the set of AMs, rules R1, R2 and R3 were applied one after the other. When comparing our final WERs with the RS component (see table \ref{tab:resultsRS}), R1 and R2 lead to an improvement of our best triphone WERs by $\unit[0.56]{\%}$. R3 slightly deteriorated our results by $\unit[0.16]{\%}$.
\par
Next, we tested different LM of orders $\{1, 2, 3, 4, 5\}$ with our best frame shift $f_{\mathrm{sh}}=\unit[12.5]{ms}$ and frame length $f_{\mathrm{len}}=\unit[20]{ms}$ configuration. Bigrams, trigrams, four-grams and five-grams performed similarly well regarding both, monophone and triphone models ($\text{WER}\approx\unit[1 \dots 2]{\%}$). We decided to stick to the trigram model which had a slight advantage with the best $\text{WER}=\unit[0.96]{\%}$. With unigrams we achieved non-comparable results since best WERs differed widely. In this case, best WERs of the monophone model ($\text{WER}=\unit[35.77]{\%}$) were also much worse than best WERs of the triphone model ($\text{WER}=\unit[18.25]{\%}$).  

\begin{table}[t!]
 \caption{Summary of the best WERs with \GRRS. We trained a GMM  (see section \ref{sec:methods_RS}) and compared the impact of different phone set minimization rules on \textit{valid} and \textit{test} sets.}
 \label{tab:resultsRS}
 \centering
 \begin{tabular}{ l c c }
  \toprule
  \multicolumn{1}{l}{\textbf{Phone Set Rule}} & \multicolumn{1}{c}{\textit{valid}} & \multicolumn{1}{c}{\textit{test}} \\
  \midrule
  --    & $0.96$ & $1.2$        \\
  +R1   & $1.05$ & $1.04$         \\
  +R1+R2   & $\bm{0.4}$ & $\bm{0.64}$         \\
  +R1+R2+R3 & $0.56$ & $\bm{0.64}$      \\
  \bottomrule
 \end{tabular}
\end{table}

\subsection{Conclusion: Read Speech}
Our ASR experiments for RS showed that the lowest WER was achieved with a lexicon with canonical pronunciation, i.e.~no pronunciation variants. The only adaptation made to this canonical lexicon was the reduction
of the phone set according to Austrian Standard German pronunciation (e.g., devoicing alveolar fricatives). State-of-the-art performance was obtained with a basic triphone model ($0.64$ WER with \textit{test}).
We observed that our methodological choices lead to large improvements (e.g., frame shift, phone set minimization, AM passes, LM orders). Additionally, the difference in WER between \textit{valid} and \textit{test} were relatively low ($\approx 0.01 - 0.24$ \%). In general, the WERs were in the range of other state-of-the art systems for RS (e.g., $1.4\%$ in \cite{Chung2021w2vBERTCC}).

\section{ASR for Conversational Speech}
\label{sec:exp2}

\begin{figure*}[ht!]
\centering
    \input{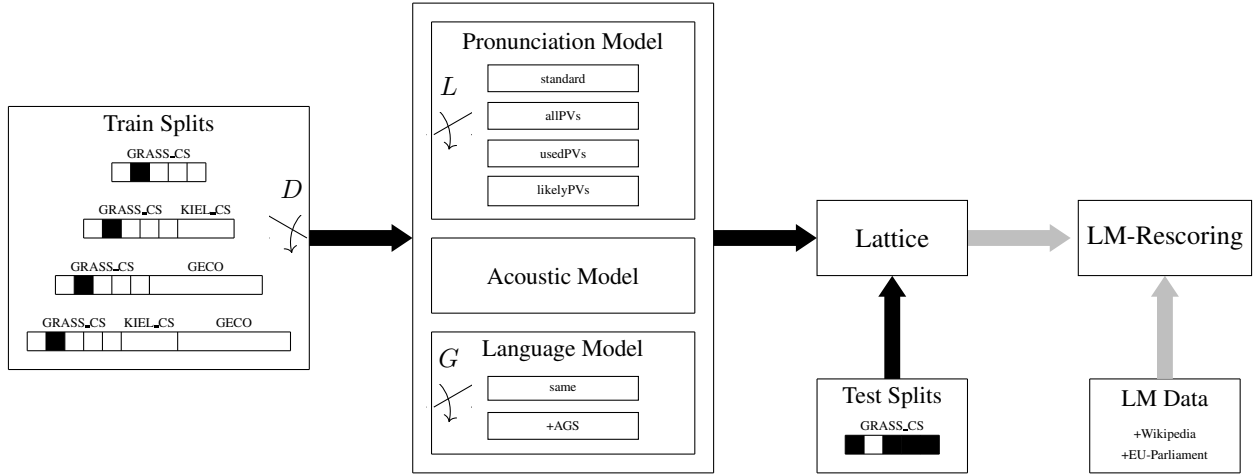}
\caption{Schematic architecture for our conversational speech recognition experiments. Training is conditioned on $4\times19$ data combinations since we introduce $4$ possible corpora combinations (switch $D$) and each conversation from \GRCS is processed individually. This results in $19$ conversation-dependent training and test splits for $4$ data combinations (see table \ref{tab:data}). Experiments analyzing the influence of the lexicon (switch $L$) utilize the same data for AM and LM training. Experiments analyzing the influence of the LM (switch $G$) utilize our standard lexicon and perform LM-rescoring as an additional option (grey arrows).}
\label{fig:flowchart}
\end{figure*}

\subsection{Methods}
\label{sec:methods_CS}
Figure \ref{fig:flowchart} shows a schematic overview of the experimental setup. AMs and LMs were trained with data from \GRCS,  GECO and \KICS (see section \ref{sec:materials}).  We present experiments which are trained merely with \GRCS, or \GRCS and GECO, or \GRCS and \KICS or \GRCS and GECO and \KICS. This study focuses on evaluating ASR on conversational Austrian German by performing leave-$p$-out cross-validation with respect to \GRCS (with $p=2$ speakers of the same conversation) resulting in approx. $0.8~$\si{\hour} of test data and $13.5~$\si{\hour} of training data per split. We randomly chose $10\%$ of resulting training splits as validation sets (approx. $1.35~$\si{\hour}) to adjust basic model parameters. When adding training data from GECO or \KICS, validation sets were built by randomly choosing $10\%$ from the newly introduced training data. For evaluation, we compared the performance on the test splits which result from the described cross-validation.
\par
In \GRCS preprocessing, we excluded chunks that contained laughter, singing, imitations/onomatopoeia, completely incomprehensible word tokens or artefacts (e.g.,~when a speaker accidentally touched their microphone). 
In case of GECO, we removed symbols indicating laughter, throat clearing and broken words from the transcriptions. In case of KICS, we removed symbols indicating laughter, smacking sounds, different types of noise and repetitions from the transcriptions. 
\par
For the AM, the ASR monophone and triphone training steps were in most parts analogous to the RS experiments. First two models were again trained with 13-dimensional MFCCs+$\Delta$+$\Delta \Delta$ and CMVN (with $f_{\mathrm{sh}}=\unit[10]{ms}$ and $f_{\mathrm{len}}=\unit[25]{ms}$). On top of the triphone GMMs (see section \ref{sec:methods_RS}), a speaker independent GMM model with linear discriminative analysis (LDA) and maximum likelihood linear transform (MLLT) \cite{Gopinath1998} was trained resulting in GMM+LDA+MLLT. Speaker-adapted training was performed also on top of GMM+LDA+MLLT with constrained maximum likelihood linear regression (fMLLR) \cite{Gales1998} resulting in GMM+fMLLR. The final triphone alignments were used to train a baseline DNN-HMM hybrid model consisting of a TDNN with $13$ layers and hidden dimensions of $512$ while utilizing only already calculated MFCC features. The network is trained with a frame-level objective function based on the cross-entropy criterion. Our recipe is based on a recipe published in \cite{GitEasyKaldi} and related DNN setups are described in \cite{rath13_interspeech, 2011vesely}. 
\par
For the LM, we used the SRILM toolkit \cite{Stolcke02} with the same configuration as in RS but trained trigrams. Here, the LM was generated given the text of all utterances from the cross-validation training splits from \GRCS and the two additional German corpora if they were also utilized for AM training. In order to evaluate a potential limited data problem when training the LMs, we also ran experiments utilizing a bigger trigram by adding approx. 220k Austrian German sentences from subtitles of broadcasts for the deaf and hard of hearing of an Austrian public television service (AGS) \cite{ORF}. 
For the latter, we performed LM-rescoring with a four-gram by again adding a random subset of 5M German sentences from AGS, German Wikipedia\footnote{\wikiurl} and the European parliament\footnote{\url{https://www.statmt.org/europarl/v7/de-en.tgz}}. In order to receive the additional LM data we adapted a toolkit which is described in \cite{Milde2018}.

\par
\subsection{Lexicon Generation} 
\label{sec:lexiconGeneration}
\par 
We created word lists from the transcriptions of all corpora. Canonical pronunciations were obtained with a G2P online tool \cite{reichel2014language}. We created four different pronunciation lexicons.
\par 
\textit{standard.} Since the German language setting of the utilized tool creates pronunciations for German Standard German (GSG)\footnote{With German Standard German, we refer to German as spoken by speakers from Germany}, we applied 6 input switch rules to obtain an Austrian Standard German pronunciation. We call the resulting pronunciation lexicon \textit{standard}. 
For the foreign language words, we changed the language setting to the corresponding language. 
\par 
\textit{allPVs.} We created a lexicon with pronunciation variants (PVs) by applying 26 phonological rules (based on findings from \cite{schuppler2014pronunciation}) to the canonical Austrian German pronunciations. 17 of these rules are relevant for conversational speech of all German varieties, e.g., assimilation of plosives and r deletion in the syllable coda, whereas 9 rules cover pronunciations that are typical for the Austrian German variety, e.g., the deletion of the syllable \enquote{ge} in the beginning of specific past participles. In addition to these rule-based variants, we also created a couple of PVs manually, in order to capture pronunciations that cannot be generated in an automated way but are frequent in Austrian German spontaneous speech, e.g., the pronunciation [\textipa{ma:}] for the word \enquote{wir} with canonical pronunciation [\textipa{vi:6}]. The resulting lexicon (\textit{allPVs}) contained on average $5.57-6.18$ variants per word.
\par 
\textit{usedPVs.} We used the \textit{allPVs} lexicon to create a forced alignment based segmentation (with a frame shift of $f_{\mathrm{sh}}=\unit[7.5]{ms}$) of all of corpora (i.e., \GRCS, \KICS and GECO). From these segmentations, we extracted the pronunciation variants that had been actually produced by the speakers, and created a pronunciation lexicon with those PVs only. The resulting lexicon (\textit{usedPVs}) had on average $1.37-1.43$ variants per word. 
\par
\textit{likelyPVs.} We created a lexicon containing only those variants which showed a high frequency of occurrence in the forced alignment, inspired by the approach presented in \cite{chen15g_interspeech}. As in \cite{chen15g_interspeech}, we calculated the statistics for the pronunciation probability estimation, but instead of integrating different probabilities for specific pronunciations, we considered only pronunciations which result in an estimated probability of $p>0.65$. This choice was made in order to give a better comparison with our other lexicons since introducing additional pronunciation probabilities to the lexicon would change the experimental design (in other words, all final lexicons involve pronunciation variants with equal probabilities by definition). The resulting lexicon (\textit{likelyPVs}) had an average of $1.16-1.26$ variants per word. Table \ref{tab:lexicons} presents a summary of all different pronunciation lexicons used in our ASR experiments.

\begin{table}[t!]
 \caption{Entry statistics of the different pronunciation lexicons used for the ASR experiments with conversational speech (see section \ref{sec:lexiconGeneration}). The number of entries is influenced by the number of utilized corpora for AM/LM training (see table \ref{tab:data} or figure \ref{fig:flowchart}).}
 \label{tab:lexicons}
 \centering
 \begin{tabular}{ l c c }
  \toprule
  \textbf{Lexicon Name} &  \textbf{$\mathrm{min}(\text{\#entries})$} & \textbf{$\mathrm{max}(\text{\#entries})$}\\
  \midrule
  standard& $13.9$k & $22.7$k        \\
  allPVs& $74.2$k & $135.1$k      \\
  usedPVs& $17.4$k & $30.4$k       \\
  likelyPVs& $14.6$k & $26.9$k    \\
  \bottomrule
 \end{tabular}
\end{table}

\begin{table*}[t!]
 \caption{Summary of conversation-dependent WERs for Austrian German conversational speech obtained with our Kaldi baseline system. The first two columns show the utilized data for AM and LM training, the third column shows the utilized lexicons and the remaining columns give mean and standard deviations of resulting $19$ WERs as well as corresponding minimum and maximum WERs.}
 \label{tab:data}
 
 \centering
 \begin{tabular}{ c m{13em} m{9em}  m{6.5em}  m{4.5em} m{3em} m{3em}}
  \toprule
   & \multicolumn{1}{l}{\textbf{AM}} & \multicolumn{1}{l}{\textbf{LM}} & \multicolumn{1}{l}{\textbf{Lexicon}} & \multicolumn{1}{l}{\textbf{\makecell{WERs}}} & \multicolumn{1}{l}{\textbf{\makecell{min(WER)}}} & \multicolumn{1}{l}{\textbf{\makecell{max(WER)}}} \\
  \midrule \midrule
   \rot{\rlap{\kern-1.5em Baseline}} & \GRCS\phantom{+GECO+\KICS} \GRCS+\KICS\phantom{+GECO} \GRCS+GECO\phantom{+\KICS} \GRCS+GECO+\KICS & same & standard & $53.89/5.18$ $53.22/5.23$ $52.58/5.59$ $\bm{51.91/5.74}$ & $42.1\phantom{/0.00}$ $42.0\phantom{/0.00}$ $40.5\phantom{/0.00}$ $\bm{39.7}\phantom{/0.00}$ & $63.7\phantom{/0.00}$ $63.2\phantom{/0.00}$ $63.9\phantom{/0.00}$ $\bm{63.2}\phantom{/0.00}$ \\ 
   \midrule \midrule
    & \GRCS\phantom{+GECO+\KICS} \GRCS+\KICS\phantom{+GECO} \GRCS+GECO\phantom{+\KICS} \GRCS+GECO+\KICS & same & \textbf{allPVs} & $55.56/5.03$ $55.16/5.21$  $54.37/5.45$ $53.62/5.56$ &  $43.9\phantom{/0.00}$   $44.4\phantom{/0.00}$ $42.6\phantom{/0.00}$ $41.6\phantom{/0.00}$ &  $64.7\phantom{/0.00}$  $64.8\phantom{/0.00}$ $64.7\phantom{/0.00}$  $64.7\phantom{/0.00}$  \\ 
   \cmidrule(lr){2-7}
   \rot{\rlap{\kern-3.5em Influence Lexicon}} & \GRCS\phantom{+GECO+\KICS} \GRCS+\KICS\phantom{+GECO} \GRCS+GECO\phantom{+\KICS} \GRCS+GECO+\KICS & same & \textbf{usedPVs} & $55.22/4.88$ $55.06/5.15$ $54.15/5.4$ $53.69/5.52$ &  $43.8\phantom{/0.00}$  $43.5\phantom{/0.00}$  $42.5\phantom{/0.00}$   $42.9\phantom{/0.00}$ &  $64.3\phantom{/0.00}$  $64.5\phantom{/0.00}$ $64.7\phantom{/0.00}$ $64.3\phantom{/0.00}$ \\ 
   \cmidrule(lr){2-7}
   & \GRCS\phantom{+GECO+\KICS} \GRCS+\KICS\phantom{+GECO} \GRCS+GECO\phantom{+\KICS} \GRCS+GECO+\KICS & same & \textbf{likelyPVs} & $ 51.87/4.88$ $51.64/5.22$ $50.93/5.61$ $\bm{50.48/5.66}$ &  $40.3\phantom{/0.00}$  $40.0\phantom{/0.00}$  $38.8\phantom{/0.00}$   $\bm{39.7}\phantom{/0.00}$ &  $\bm{60.9}\phantom{/0.00}$  $61.9\phantom{/0.00}$ $62.2\phantom{/0.00}$ $62.0\phantom{/0.00}$ \\ 
  \midrule \midrule
   & \GRCS\phantom{+GECO+\KICS} \GRCS+\KICS\phantom{+GECO} \GRCS+GECO\phantom{+\KICS} \GRCS+GECO+\KICS & \textbf{+AGS (220k)} & standard & $52.26/5.5$ $51.53/5.58$ $51.38/5.72$  $50.74/5.82$ &  $40.1\phantom{/0.00}$  $40.3\phantom{/0.00}$ $40.3\phantom{/0.00}$  $39.0\phantom{/0.00}$ &  $62.9\phantom{/0.00}$  $62.8\phantom{/0.00}$  $63.3\phantom{/0.00}$ $62.3\phantom{/0.00}$ \\ 
   \cmidrule(lr){2-7}
   \rot{\rlap{\kern-0em Influence LM}} & \GRCS\phantom{+GECO+\KICS} \GRCS+\KICS\phantom{+GECO} \GRCS+GECO\phantom{+\KICS} \GRCS+GECO+\KICS & \textbf{+AGS (220k)\phantom{rescore} +Rescoring (5M)} & standard & $51.17/5.26$ $50.91/5.65$ $50.92/5.82$ $\bm{50.19/5.82}$ &  $39.5\phantom{/0.00}$ $39.8\phantom{/0.00}$  $39.5\phantom{/0.00}$  $\bm{38.7}\phantom{/0.00}$ &  $62.4\phantom{/0.00}$  $62.4\phantom{/0.00}$ $63.2\phantom{/0.00}$ $\bm{62.2}\phantom{/0.00}$ \\ 
   \midrule \midrule
   \rot{\rlap{\kern-0.75em Best}} & \GRCS\phantom{+GECO+\KICS} \GRCS+\KICS\phantom{+GECO} \GRCS+GECO\phantom{+\KICS} \GRCS+GECO+\KICS & \textbf{+AGS (220k)\phantom{rescore} +Rescoring (5M)} & \textbf{likelyPVs} &  $49.15/5.28$ $49.02/5.69$ $49.07/5.88$ $\bm{48.5/6.09}$ &  $37.3\phantom{/0.00}$  $37.3\phantom{/0.00}$  $37.2\phantom{/0.00}$   $\bm{37.0}\phantom{/0.00}$ &  $\bm{59.1}\phantom{/0.00}$  $60.4\phantom{/0.00}$ $60.8\phantom{/0.00}$ $61.3\phantom{/0.00}$  \\ 
   \bottomrule
 \end{tabular}
 \label{tab:resultsCS}
\end{table*}

\subsection{Results}
Table \ref{tab:resultsCS} shows the ASR results for different training setups, always using \GRCS as test data. We compared ASR experiments with (1) different data sizes for AM training, (2) lexicons of different amounts of variants (3) LMs trained on different data sizes and (4) a combination of the best AM, Lexicon and LM.
With respect to the Acoustic Model (AM), all AMs showed benefits from additional training data from other corpora, resulting in absolute WER improvements of approx.~$1-2\%$ with respect to mean values; then again, respective standard deviations are higher when more data is used indicating that overall performance improves but robustness problems arise. 
With respect to the pronunciation lexicon, our results showed that in comparison to using our standard lexicon, lexicons with very high numbers of variants (i.e., \textit{allPVs} and \textit{usedPVs}) lead to a performance decrease. With the \textit{likelyPVs} lexicon, however, which contained a small number of likely pronunciation variants, performance improved by approx. $1.5\%$ compared to the best mean value of the baseline with the standard lexicon. 
With respect to varying the amount of training data for the LM, we achieved the best results by adding data from all corpora for AM training, adding data from all corpora plus AGS for LM training, a lexicon with likely pronunciation variants and LM-rescoring with our 5M additional German sentences (see section \ref{sec:materials}) resulting in a best mean WER of $48.5\%$. 
\par
Overall, when comparing the best mean WER with our baseline system, we achieved an absolute WER improvement of approx. $4.5\%$. In general, in all experiments, we observed highly varying WERs between the different conversations (i.e., speaker pairs) of \GRCS (standard deviations range from $4.88\%$ to $6.09\%$).


\subsection{Discussion}
\label{sec:discussion}

\par
This paper aimed at building a Kaldi-based ASR system for Austrian German, with a focus on conversational speech. Since our first experiments already showed large differences from speaker pair to speaker pair, we decided to provide cross-validation results in order to get more insight into conversation-dependency of ASR systems. It is worth noting that, even though when reaching performance gains by certain methodological choices, we still observed similarly high standard deviations of the WERs across the conversations. Hence, neither the change of data sizes for AM and LM training nor the different approaches for pronunciation modeling made the ASR system more robust to variation in conversational speech \cite{linke-etal-2022-conversational}.
\par
In comparison to other benchmarks, our results form the cross-validation approach highlights how challenging the task of conversational speech recognition is.  Other benchmarks tend to train and test on pre-defined training and test sets which may cause an optimistic bias towards ASR accuracy \cite{szymanski-etal-2020-wer}. The cross-dialect analysis described in \cite{Elfeky2018}, for instance, showed how ASR performance decreases when a dialect variation is evaluated on a system which had been trained on another dialect of the same language. We hypothesize that testing each conversation individually shows a similar effect because even though speakers in GRASS had a comparable regional background, we still find high individual dialectical variation which is in line with a previous analysis of the corpus in \cite{schuppler2014pronunciation}. 
Our results suggest not only to investigate how to improve overall ASR performance but to focus more of tackling missing robustness, especially in case of conversational speech recognition.
\par
ASR with our standard lexicons and a large amount of additional LM data resulted in a mean WER of $50.19\%$. When utilizing a lexicon with likely pronunciation variants only (by adding approx. $4.2$k entries to the standard lexicon; see table \ref{tab:lexicons}) without adding a large amount of additional LM data, we achieved a mean WER of $50.48\%$. Thus, when comparing our results from using different lexicons with those from using different amounts of data for the LM (cf., table \ref{tab:resultsCS}), we observe that training LMs with large amounts of data had a similarly high impact on improving WERs (approx.~1.5\%) as using the best pronunciation lexicon (i.e., \textit{likelyPVs}). A survey on modeling pronunciation variation for ASR \cite{STRIK1999225} summarizes that adding pronunciation variants to the lexicon appears to improve recognition performance especially if the different frequencies of occurrence of variants are considered. Two decades and many ASR architectures later, our results still confirm their observation. We further showed that well-developed pronunciation modeling, for which no additional data resources nor high computational efforts are necessary, could compensate for the necessity of collecting more LM data. Yet, the combination of both methodological approaches (pronunciation modeling and collecting more LM data) still yielded the best results. This finding is especially relevant for the field of low-resource ASR.

\section{Conclusion}
This paper presented the development of an ASR system for a corpus of read (RS) and conversational (CS) Austrian German, dealing with two challenges: (1) the conversational speech recorded is highly casual and (2) Austrian German is a variety of German for which little speech data is available in general, but even more so for conversational speech. For RS, we achieved the best results with $f_{\mathrm{sh}}=12.5\si{\milli\second}$, a canonical lexicon, a reduced phone set and LM orders greater than $N=2$. For our baseline in CS, we achieved best results with a hybrid DNN-HMM model, a lexicon including the most likely pronunciation variants and an LM trained on a combination of spontaneous German and Austrian German. 
We achieved similar performance gains by either incorporating knowledge into the pronunciation lexicon or augmenting the training data. We observed high variation in performance from conversation to conversation (i.e., approx.~$\unit[5\!-\!6]{\%}$ standard deviation), regardless of the overall performance, indicating low robustness of the ASR system for conversational speech. The reasons for the lack of robustness could come from high variation with respect to pronunciation variation (i.e., dialectal background), speech rate variation (in CS speech rate varies from $0.88$ to $45.45$ phones per second with a mean of $12.38$ and a standard deviation of $4.28$), differences in lexical choice (as the topics are chosen freely), differences with respect to whether complete syntactic structures are used by the speakers and their turn-taking behaviour. In future work, we plan to analyze in detail which are the factors that hinder robust ASR of conversational speech. 

\section{Acknowledgements}
\thanks{The work by Saskia Wepner was funded by grant P-32700-N from the Austrian Science Fund (FWF).}

\bibliographystyle{IEEEbib}
\bibliography{strings,refs}

\end{document}